\documentclass[conference]{IEEEtran}
\IEEEoverridecommandlockouts

\usepackage{cite}
\usepackage{amsmath,amssymb,amsfonts}
\usepackage{algorithmic}
\usepackage{graphicx}
\usepackage{textcomp}
\usepackage{xcolor}
\usepackage{hyperref}
\usepackage[linesnumbered,ruled,vlined]{algorithm2e}
\usepackage{subfig}
\def\BibTeX{{\rm B\kern-.05em{\sc i\kern-.025em b}\kern-.08em
    T\kern-.1667em\lower.7ex\hbox{E}\kern-.125emX}}
\begin{document}

\title{Robustifying 3D Perception via Least-Squares Graphs for Multi-Agent Object Tracking
\thanks{This work has received funding from the EU’s Horizon Europe research and innovation programme in the frame of the CoEvolution project “A comprehensive trustworthy framework for connected machine learning and secured interconnected AI solutions” under the Grant Agreement No 101168560.
}
}
\author{
\IEEEauthorblockN{Maria Damanaki$^{1,3}$, Ioulia Kapsali$^{1}$, Nikos Piperigkos$^{1,2}$, Alexandros Gkillas$^{1,2}$, Aris S. Lalos$^{1,2}$}
\IEEEauthorblockA{$^1$Industrial Systems Institute, Athena Research Center, Patras Science Park, Greece\\
$^2$AviSense.AI, Patras Science Park, Greece, $^3$ Dpt. of Informatics \& Telecom., University of Ioannina, Arta, Greece\\
Emails:\{mdamanaki, ikapsali\}@isi.gr, \{piperigkos, gkillas\}@avisense.ai, lalos@athenarc.gr 
}
}
\maketitle

\begin{abstract} 
The critical perception capabilities of EdgeAI systems, such as autonomous vehicles, are required to be resilient against adversarial threats, by enabling accurate identification and localization of multiple objects in the scene over time, mitigating their impact. 
Single-agent tracking offers resilience to adversarial attacks but lacks situational awareness, underscoring the need for multi-agent cooperation to enhance context understanding and robustness.
This paper proposes a novel mitigation framework on 3D LiDAR scene against adversarial noise by tracking objects based on least-squares graph on multi-agent adversarial bounding boxes. 
Specifically, we employ the least-squares graph tool to reduce the induced positional error of each detection's centroid utilizing overlapped bounding boxes on a fully connected graph via differential coordinates and anchor points.  
Hence, the multi-vehicle detections are fused and refined mitigating the adversarial impact, and associated with existing tracks in two stages performing tracking to further suppress the adversarial threat. 
An extensive evaluation study on the real-world V2V4Real dataset demonstrates that the proposed method significantly outperforms both state-of-the-art single and multi-agent tracking frameworks by up to \textbf{23.3\%} under challenging adversarial conditions, operating as a resilient approach without relying on additional defense mechanisms.
\end{abstract}

\section{Introduction}
\label{intro}
Autonomous driving forms a crucial core of EdgeAI systems demanding accurate and robust perception to ensure efficiency and safety on the roads. Single-agent perception is susceptible to sensor failures, environmental conditions, limited sensing range, and malicious attacks. Contrary, multi-agency ensures high situational awareness via the exchange of contextual information among Connected and Autonomous Vehicles (CAVs). 
However, adversarial attacks can undermine the credibility, accuracy and safety of Single-Agent Multi-Object Detection (SAOD) and Tracking (SAMOT), as well as Multi-Agent Multi-Object Detection (MAOD) and Tracking (MAMOT), which form the core of EdgeAI perception by detecting, localizing and identifying objects over time.  


State-of-the-art trackers typically consist Deep Neural Networks (DNNs) in tracking-by-detection or jointly detection and tracking paradigms, making them vulnerable to adversarial threats. These attacks involve small imperceptible perturbations that alter the expected output \cite{szegedy2013intriguing}, degrading perception performance. Several works are focused on generating adversarial attacks by perturbing, adding or dropping points \cite{xiang2019generating,liu2020adversarial,wicker2019robustness,zhang2024comprehensive} in the 3D scene.
Moreover, spatiotemporal adversarial perturbations or patches in \cite{jia2024robust,zhang2024dual,jia2020fooling, long2024papmot}, deteriorate object detection and cause identity errors across frames.
AttackZone \cite{muller2022physical} physically hijacks Siamese-based trackers\cite{shuai2021siammot}. 
A cooperative attack \cite{tu2021adversarial} injects perturbations into shared DNN representations, degrading perception. 
Mitigating adversarial effects requires the development of robust defense schemes. Adversarial training \cite{zhang2024comprehensive, wu2024enhancing} is a widely adopted defense approach, where DNNs are trained with adversarial examples to improve their resilience. Moreover, input transformations such as point cloud rotation, flipping, Gaussian noise injection, and point quantization 
\cite{zhang2024comprehensive},\cite{yang2019adversarial} operate as resilient methods. In \cite{jia2024robust}, adversarial perturbations are estimated in order to subtract their impact. Cooperative defense approaches as MADE \cite{zhao2024made} 
detect and suppress malicious agents, while methods as in \cite{ding2023robust}, agents exchange critical features optimized via mutual information constraints.
Nevertheless, all the above methods are computationally expensive, either due to the need of learning the characteristics of adversarial perturbations prior to mitigation or to perform extensive cross-agent verification.
Therefore, we propose a novel and cost-efficient multi-agent mitigation framework that leverages shared information among agents to reduce adversarial perturbations operating as a denoiser, without training, learning parameters or additional defense mechanisms. More specifically, we formulate a fully connected graph topology based on the perturbed predicted bounding boxes and we apply least-squares graph optimization methods inspired by least-square meshes \cite{sorkine2004least}. The proposed Graph Signal Processing method integrates multi-vehicle adversarial detections by exploiting overlapped information among agents and mitigates the induced error on their centroids usind both estimated differential coordinates and estimated anchor points. 
Thereafter, existing trajectories are associated in two-stages with the resilient detections, updated via a linear Kalman Filter, forming the tracking pipeline of our novel method, and addressing misdetections (i.e. detections of nonexistent objects and missed detections of real objects) suppressing further the impact of attacks. We conduct extensive evaluations on the real-world V2V4Real dataset and benchmarking our method against other state-of-the-art SAMOT and MAMOT approaches as well as with a classical defense mechanism to highlight our method's superiority. To sum up, the main contributions of this study can be summarized as follows: 
\begin{itemize}
\item We introduce a novel graph signal processing based MAMOT concept as a robust adversarial mechanism, without relying on additional defenses, enabling resilient 3D perception in CAVs.
\item
A denoising mitigation approach is derived to reduce adversarial perturbations exploiting overlapped multi-agent adversarial detections using differential coordinates and estimated anchors followed by a two-stage tracking association procedure for further suppression robustifying situational awareness.
\item
We extensively evaluate our novel framework on the real-world V2V4Real \cite{10203124} against state-of-the-art MAMOT and SAMOT approaches considering attacks either to ego or both vehicles and demonstrating promising results on tracking accuracies up to \textbf{23.3\%}. 
\end{itemize}

\section{Preliminaries}
This Section introduces the measurement and system models of our framework, as well as the type of adversarial attacks that will be considered.
\label{preliminaries}
\subsection{Measurement and System Models}
\label{systemmodel}
Our MAMOT framework follows the tracking-by-detection paradigm, where each agent's detections are projected into a common coordinate system in order to leverage multi-vehicle data fusion and tracking, and thus mitigate adversarial attacks. Specifically, at each time step $t$, each agent $i$ acquires a set of $R_{D_i}$ detections denoted by $\mathcal{D}_i^t = {\{ \boldsymbol{x_{D}^{(i,m,t)}} \in \mathbb{R}^7 \mid m = 1, 2, \ldots, R_{D_i} \}} \in \mathbb{R}^{R_{D_i} \times 7}$,
with $\boldsymbol{x_{D}^{(i,m,t)}} =[x_{i,m} \ y_{i,m} \ z_{i,m} \ \theta_{i,m} \ h_{i,m} \ w_{i,m} \ l_{i,m}]^T \in \mathbb{R}^7$ representing the $m$-th detection. Here, $(x_{i,m}, y_{i,m}, z_{i,m})$ denotes the centroid of the 3D bounding box, $\theta_{i,m}$ is the yaw angle around the $z$-axis, and $h_{i,m}, w_{i,m}, l_{i,m}$ correspond to the height, width, and length of the bounding box, respectively.
The set of $R_O$ real objects of agent $i$ at time instant $t$ is denoted by ${\mathcal{B}_i^t} \in \mathbb{R}^{R_O \times 7}$ where each object is described by its centroid, yaw angle and dimensions, similarly to the set $\mathcal{D}_i^t$.
Furthermore, the set of $R_I$ tracked objects at time instance $t$ is described as 
$\mathcal{I}^t = {\{\boldsymbol{x_I^{(r,t)}} \in \mathbb{R}^{10} \mid r = 1, 2, \ldots, R_I \}} \in \mathbb{R}^{R_I \times 10}$ with the r-th tracked object be $\boldsymbol{x_{I}^{(r,t)}}=[x_r \ y_r \ z_r \ \theta_r \ h_r \ w_r \ l_r \ u_{x_r} \ u_{y_r} \ u_{z_r}]^T  \in \mathbb{R}^{10}$, where $(x_r,y_r,z_r)$ represents its centroid, $\theta_r$ the angle around z-axis, $h_r,w_r,l_r$ the height, width, length, and $(u_{x_r}, u_{y_r}, u_{z_r})$ the 3D linear velocity, respectively. Note that the tracked object is the expected output of our resilient framework and it is not directly related with a specific agent contrary to the agents' detections. 
Moreover, we employ a state transition and measurement model to perform tracking using linear Kalman Filtering: 
\begin{itemize} 
\item State transition model: 
\begin{align}
    \begin{split}
    \label{CV}
    \boldsymbol{x_{I}^{(r,t)}} = f(\boldsymbol{x_{I}^{(r,t-1)}},\boldsymbol{v_{x}^{(r,t)}}), \boldsymbol{\eta_{x}^{(r,t)}} \sim \mathcal{G}(0, \boldsymbol{\Sigma_{\eta_{x}}})
    \end{split}
\end{align}
\end{itemize}
\begin{itemize}
    \item Measurement model:
    \begin{align}
    \label{self_positioning}
        \begin{split} \boldsymbol{z_{I}^{(r,t)}} = \boldsymbol{x_{I}^{(r,t)}} + \boldsymbol{\eta_z^{(r,t)}},  \boldsymbol{\eta_z^{(r,t)}} \sim \mathcal{G}(0, \boldsymbol{\Sigma_{\eta_{z}}})
        \end{split}
    \end{align}
\end{itemize}
The state transition function $f(\cdot)$ is defined by a constant velocity model, while both models are degraded by additive white Gaussian noise $\boldsymbol{\eta_{x}^{(r,t)}}$, and $\boldsymbol{\eta_z^{(r,t)}}$, respectively. According to linear Kalman Filter, each track's state can be defined by the prediction and update equations as follows:
\begin{itemize}
    \item State Prediction:
    \begin{align}
        \boldsymbol{\bar{x}_{I}^{(r,t)}} &= \boldsymbol{F^t}\boldsymbol{\hat{x}_{I}^{(r,t-1)}} + \boldsymbol{\eta_{x}^{(r,t)}} \label{eq:prediction_sta}\\
        \boldsymbol{{\bar{P}^{(r,t)}}} &= \boldsymbol{F^t\hat{P}^{(r,t-1)}(F^t)^{T} + Q^{(r,t)}} \label{eq:prediction_P}
    \end{align}
    \item State Update:
    \begin{align}
    \label{State Update}
        \boldsymbol{K^{(r,t)}} &=\boldsymbol{{\bar{P}^{(r,t)}}(H^t)^{T} [H^t{\bar{P}^{(r,t)}}{(H^t)}^{T} + R^{(r,t)}]^{-1}} \\
        \boldsymbol{\hat{x}_{I}^{(r,t)}} &= \boldsymbol{\bar{x}_{I}^{(r,t)}} + \boldsymbol{K^{(r,t)}[\boldsymbol{z_{I}^{(r,t)}} - H^t \boldsymbol{\bar{x}_{I}^{(r,t)}}] }\label{eq:upd_x} \\
        \boldsymbol{\hat{P}^{(r,t)}} &= \boldsymbol{ {\bar{P}^{(r,t)}} - K^{(r,t)}H^t{\bar{P}^{(r,t)}}} \label{eq:upd_P}
    \end{align}
\end{itemize}
where $\boldsymbol{F^t}\in \mathbb{R}^{10\times10}$ is the state transition matrix, $\boldsymbol{P^{(r,t)}}\in \mathbb{R}^{10\times10}$ the covariance matrix of the tracked object, $\boldsymbol{Q^{(r,t)}}\in \mathbb{R}^{10\times10}$ the process noise covariance matrix, 
$\boldsymbol{K^{(r,t)}}\in \mathbb{R}^{10\times7}$ the Kalman gain, $\boldsymbol{H^t}\in \mathbb{R}^{7\times10}$ the measurement model matrix, and $\boldsymbol{R^{(r,t)}}\in \mathbb{R}^{10\times10}$ the measurement noise covariance matrix.
Finally, the set $\mathcal{I}^t = \{\boldsymbol{\hat{x}_{I}^{(r,t)}}|r=1, \ldots, R_I\}\in \mathbb{R}^{R_I\times 10}$ contains the state of tracked objects, after performing Kalman Filtering.
Therefore, the set of detected and tracked bounding boxes will serve as the foundation to formulate our framework.

\subsection{Adversarial Perturbation Attack}
\label{attack}
\begin{figure} 
\centering
 \includegraphics[scale=0.73]{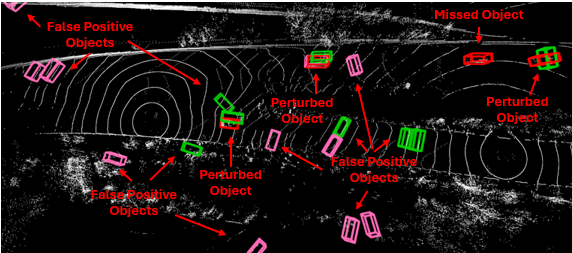}
  \caption{Impact of the adversarial perturbation attacks on two CAVs result in increased false detections, perturbed predicted bounding boxes, and not detected (missed) objects, and thus degrading detection and overall perception performance (pink and green: detections from two agents, red: ground-truth objects).}
  \label{fig:adversarial_attack}
\end{figure}

To validate the resiliency of our method, we design an untargeted point-wise perturbation attack \cite{zhang2024comprehensive} on the agents' point clouds. 
This attack, as described in \cite{zhang2024comprehensive}, introduces imperceptible perturbations to the point clouds, causing the 3D detector to infer misdetections, and spatially displaced detections from ground truth locations, as also illustrated in Fig\ref{fig:adversarial_attack}, deteriorating the detection and overall performance. 
Specifically, at time step $t$, structured modifications are introduced via noise vectors $\boldsymbol{g_i^t} \in \mathbb{R}^{\Pi_i\times 3}$ to the coordinates $x,y,z$ of agent $i$'s benign point cloud $\boldsymbol{Y_i^t} \in \mathbb{R}^{\Pi_i \times 3}$ where $\Pi_i$ is the number of points.
At time instance $t$, the adversarial point cloud of agent $i$ is described by $\boldsymbol{A_i^t}=\boldsymbol{Y_i^t}+\boldsymbol{g_{i}^t} \in \mathbb{R}^{\Pi_i\times 3}$.
To perform the attack, the problem is formulated as a gradient-based optimization algorithm with a dual loss function: $\min_{\boldsymbol{A_i^t}} \mathcal{L}(\boldsymbol{A_i^{t}}, \mathcal{B}_i^t) = \lambda \mathcal{L}_{\text{pert}}(\boldsymbol{A_i^{t}}, \boldsymbol{Y_i^t}) - \mathcal{L}_{\text{det}}(\boldsymbol{A_i^{t}}, \mathcal{B}_i^t)$, where  $\mathcal{L}_{\text{pert}}(\boldsymbol{A_i^{t}}, \boldsymbol{Y_i^t})$ is the $\ell_2$-norm of the perturbation constraining the magnitude of $\boldsymbol{g_i^t}$,  $\mathcal{L}_{\text{det}}(\boldsymbol{A_i^{t}},\mathcal{B}_i^t)$ is the detection loss used by the 3D detector, 
$\lambda$ balances the trade-off between the two terms. At $k$ iteration, the adversarial point cloud is updated via: 
\begin{align}
 \boldsymbol{A_{i}^{(k,t)}} = \text{Clip}_{\boldsymbol{Y_{i}^{t}}, \varepsilon} \left( \boldsymbol{A_{i}^{(k-1,t)}} - \alpha \cdot \frac{\nabla_{\boldsymbol{A_{i}^{(k-1,t)}}} \mathcal{L}}{ \| \nabla_{\boldsymbol{A_{i}^{(k-1,t)}}} \mathcal{L} \|_2 } \right)
    \label{point_clouds_after}
\end{align}
where $\boldsymbol{A_{i}^{(0,t)}} = \boldsymbol{Y_i^t}$, and 
$\varepsilon$ is the maximum allowed per-point displacement, enforcing a bounded distortion constraint relative to the original point cloud through the clipping function.
Hence, the adversarial attack causes displacements on bounding boxes relative to the real object locations along with misdetections degrading detection and overall performance. Thus, the design of denoising approaches are essential to mitigate such attacks. Therefore, we propose a novel resilient method to reduce the noise of the displaced detections via least-squares graph optimization and tracking and association procedures to address misdetections, robustifying perception without any training module.

\section{Adversarial Resilient Least-squares Tracking}
\label{ARLOT}
In this Section, the proposed method for resilient perception will be introduced, relying on the least-squares graph scheme for effectively smoothing multi-agent adversarial 3D bounding boxes via a graph-aware topology optimization.
\subsection{Least-Squares Graph Scheme}
\label{sec:lsm}
Our novel resilient framework employs the least-squares graph to optimally reduce the induced positional errors of multi-agent adversarial bounding boxes centroids, and thus mitigate adversarial attacks. This technique is a well-established tool \cite{sorkine2004least} for smoothing vertices positions in 3D graph via differential coordinates where each vertex lies in the center of gravity of its neighbors in least-squares sense. 
More specifically, at time step $t$ consider an undirected graph with $N^t$ nodes represented as $O^{t} = (\mathcal{U}^t, \mathcal{E}^t)$, where $\mathcal{U}^t$ and $\mathcal{E}^t$ denote the set of vertices and edges, and respectively. The system is formulated as $\boldsymbol{L^tu^t}=\boldsymbol{\delta^t}$, where $\boldsymbol{L^t} \in \mathbb{R}^{N^t \times N^t}$ is the Laplacian matrix which encodes the connections among vertices,  $\boldsymbol{u^t} \in \mathbb{R}^{N^t}$ is the vector of nodes, and $\boldsymbol{\delta^t} \in \mathbb{R}^{N^t}$ are the differential coordinates which capture the inherent geometries of nodes.
Note the Laplacian matrix is equal to $\boldsymbol{L}^t = \boldsymbol{D}^t - \boldsymbol{S}^t$, where $\boldsymbol{D}^t \in \mathbb{R}^{N^t\times N^t}$ is the diagonal degree matrix with entries indicating the degree of connected nodes, $\boldsymbol{S^t} \in \mathbb{R}^{N^t \times N^t}$ is the adjacency matrix with 0 in diagonal and -1 to the adjacent nodes, 
$\boldsymbol{u^{t}} = [u^{(1,t)} \ u^{(2,t)} \ \hdots \ u^{(N^t,t)}]$ where $u^{(n,t)}$ is the spatial attribute of each $n$ node and $\boldsymbol{\delta^{t}} = [\delta^{(1,t)} \  \delta^{(2,t)} \ \hdots \ \delta^{(N^t,t)}]$,  
where $\delta^{(n,t)}= \sum_{v}^{{N^t}}({u^{(n,t)}}-{u^{(v,t)}})$ is the differential coordinate of each $n$ node with respect to its $v$ neighboring vertices. 
Due to the singularity of the Laplacian matrix, Laplacian is extended to $\boldsymbol{\tilde{L}}^{t} = \begin{bmatrix} \boldsymbol{L}^{t} \\ \boldsymbol{\mathbb{I}}_{N^t}^{t} \end{bmatrix} \in \mathbb{R}^{2N^t \times N^t}$ incorporating the identity matrix $\mathbb{I}_{N^t}^t \in \mathbb{R}^{N^t\times N^t}$, and the differential vector is reconstructed to $\boldsymbol{\tilde{\delta^t}} = [\boldsymbol{\delta^t \ c^t}]^T \in \mathbb{R}^{2N^t}$ by adding anchor vertices $\boldsymbol{c^{t}} = [c^{(1,t)} \  c^{(2,t)} \ \hdots \ c^{(N^t,t)}]^T \in \mathbb{R}^{N^t}$ with $c^{(n,t)}$ the additional information of the vertex $u^{(n,t)}$.
Hence, in each spatial attribute $x,y,z$, we solve the least-squares problem:
\begin{align}
    \boldsymbol{u}^t = \arg\min_{\boldsymbol{u}^t} \left\| \boldsymbol{\tilde{L}}^t \boldsymbol{u}^t - \boldsymbol{\tilde{\delta}}^t \right\|^2
    \label{eq:Minimization}
\end{align}
with the unique analytical solution:
\begin{equation}
    \boldsymbol{u}^t = \left( (\boldsymbol{\tilde{L}^t})^T \boldsymbol{\tilde{L}^t} \right)^{-1} (\boldsymbol{\tilde{L}^t})^T \boldsymbol{\tilde{\delta}}^t
    \label{eq:LSMsol}
\end{equation}

\subsection{Adversarial Resilient Least-squares Object Tracking Framework}

The proposed \textbf{Adversarial Resilient Least-squares Object Tracking (ARLOT)} method lays on the least-squares graph formulation to mitigate adversarial perturbations on multi-vehicle detections and two stages of tracking association to address the increased number of misdetections suppressing further the adversarial impact and robustifying 3D perception.
More specifically, the \textbf{ARLOT} leverages the overlapping contextual information of CAVs to design a fully connected graph over the multi-agent adversarial bounding boxes, and denoising their centroids from the induced errors in least-squares sense. Thereafter, performs robust tracking in two stages by updating trajectories states via the resilient multi-agent detections, handling misdetections and effectively mitigating the adversarial impact. 
The proposed method will be derived assuming two CAVs, but it is easily scalable to a larger number of agents. 
Firstly, adversarial point clouds are generated by Eq.\ref{point_clouds_after}, depending on whether one or both CAVs are attacked as described in Section \ref{attack}, causing misdetections and displaced detections on the inference of each agent's 3D object detector.
When agents observe the same object, the shared detections introduce overlapping information. As such, multi-agent perturbed detections are associated using the Hungarian Algorithm (HA) and the 3D Intersection over Union (IoU) as a similarity metric.
The set of $m_i$ overlapped detections and $u_i$ non-overlapped of the $i$ are described by $m\mathcal{D}_i^t \in \mathcal{R}^{m_i\times 7}$, and $u\mathcal{D}_i^t \in \mathcal{R}^{u_i\times 7}$, respectively, and similar quantities are denoted for agent $j$. 
To effectively model and exploit these interactions, we consider an undirected fully connected graph ${O}^{t} = (\mathcal{U}^t, \mathcal{E}^t)$, where the set of nodes $\mathcal{U}^t=\{\mathcal{D}_i^{t},\mathcal{D}_j^{t}\}$ and $\mathcal{E}^t$ the connections among them and each edge is equal to 1. At time instant $t$, the adversarial detections $\mathcal{D}_i^t$ from agent $i$ are interconnected among them, and also linked to the adversarial detections $\mathcal{D}_j^t$ from agent $j$. Note that we focus on estimating only the 3D centroid of each bounding box. To leverage the overlapping information, we define two anchor vectors $\boldsymbol{c_{ij}^{t}}=
    \begin{bmatrix}
     m\boldsymbol{x_{j,m}} & m\boldsymbol{x_{j,m}}& u\boldsymbol{x_{i,m}} & u\boldsymbol{x_{j,m}}
    \end{bmatrix}^T$,
$\boldsymbol{c_{ji}^{t}} =
    \begin{bmatrix}
     m\boldsymbol{x_{i,m}} & m\boldsymbol{x_{i,m}} & u\boldsymbol{x_{i,m}} & u\boldsymbol{x_{j,m}}
    \end{bmatrix}^T \in \mathbb{R}^{N^t}$ containing the complimentary information of x-part of $\{\boldsymbol{x_D^{(i,m,t)}}, \boldsymbol{x_D^{(j,n,t)}}\}$ in case of two CAVs. 
Moreover, the extended Laplacian matrix which captures connections among perturbed detections and the differential vector which contains all the available information are calculated according to \ref{sec:lsm}. Hence, we solve the system of Eq.\ref{eq:Minimization} and estimate two sets of analytical solutions from the Eq. \ref{eq:LSMsol} for the x-part of each detection accordingly: 
\begin{align}
    \boldsymbol{J_{ij}^{t}} &= \left( \left(\tilde{\boldsymbol{L}}^t\right)^T \tilde{\boldsymbol{L}}^t \right)^{-1} \left( \tilde{\boldsymbol{L}}^t \right)^T \left[\boldsymbol{\delta^t} \ \boldsymbol{c_{ij}^t}\right]^T, \boldsymbol{J_{ij}}^{t} \in \mathbb{R}^{N^t}
    \label{eq:LSMsol1}
\end{align}
\begin{align}
    \boldsymbol{J_{ji}^{t}} &= \left( \left(\tilde{\boldsymbol{L}}^t\right)^T \tilde{\boldsymbol{L}}^t \right)^{-1} \left( \tilde{\boldsymbol{L}}^t \right)^T \left[\boldsymbol{\delta^t} \ \boldsymbol{c_{ji}^t}\right]^T,  \boldsymbol{J_{ji}}^{t} \in \mathbb{R}^{N^t}
    \label{eq:LSMsol2}
\end{align}
where the resilient and fused detections corresponding to the anchor vector $\boldsymbol{c_{ij}^t}$ are represented as $\boldsymbol{J_{ij}^t} \in \mathbb{R}^{N^t}$ and similarly to the $\boldsymbol{c_{ji}^t}$ as $\boldsymbol{J_{ji}^t} \in \mathbb{R}^{N^t}$. Similar equations are followed for $y$ and $z$ attributes, and thus the two sets of resilient detections are $\mathcal{J}_{ij}^t \in \mathbb{R}^{N^t \times 7}$ and $\mathcal{J}_{ji}^t \in \mathbb{R}^{N^t \times 7}$.

After mitigating the perturbation error of each multi-agent adversarial detection utilizing overlapped information via least-squares graph, the tracking procedure takes place to address misdetections. To be more detailed, the set $\mathcal{J}_{ij}^t$ is associated with the existing tracks $\mathcal{I}^t$ using HA and 3D IoU. Upon successful associations, $m\mathcal{I}^t$ tracks update their states with the resilient associated detections $m\mathcal{J}_{ij}^t$ by Eq.\ref{eq:upd_x},\ref{eq:upd_P}. In case of unmatched refined detections $u\mathcal{J}_{ij}^t$, new tracks are initialized. 
Moreover, the unmatched tracks $u\mathcal{I}^t$ are associated with the $\mathcal{J}_{ji}^t$ resilient detections employing once again the HA and the 3D IoU. Hence, $m\mathcal{I}^t, m\mathcal{J}_{ji}^t$, represent the matched tracks and optimized detections and $u\mathcal{I}^t$ and $u\mathcal{J}_{ji}^t$ the unmatched tracks and unmatched detections. This process continues based on the number of agents. 
Thereafter, a lifetime management module is handling each track's state \cite{9341164}. More specifically, a track is "confirmed", if it achieves successful associations over a predefined number of consecutive time steps, denoted as $hits$. Conversely, if a track fails to establish an association for a specified number of consecutive time steps, noted as $age$, it is considered as "dead" and is removed from further processing. 
At the next time step $t+1$, all active tracks are predicted by Eq.({\ref{eq:prediction_sta}},{\ref{eq:prediction_P}}).
Moreover, the computational complexity relies on the least-squares graph with complexity equal or lower than $\mathcal{O}((2N^t N^t)^2)$ \cite{piperigkos2024graph}. 
Therefore, the \textbf{ARLOT} mitigates positional errors of multi-agent adversarial detections and updates trajectories states through two-stages association with smoothed detections, ensuring resilience without relying on other defense mechanisms. The proposed framework is summarized in \textbf{Algorithm \ref{algor:ARLOT}}.

\begin{algorithm}[ht]
\resizebox{0.9\linewidth}{!}{
\begin{minipage}{\linewidth}
\small
\textbf{Input}: Adversarial detections $\mathcal{D}_i^t$ (agent $i$), $\mathcal{D}_j^t$ (agent $j$), Tracks $\mathcal{I}^t$ \\
\textbf{Output}: Tracks $\mathcal{I}^{t+1}$ \\
\For {each timestep $t = 1, 2, \ldots, T$}{
    Formulate fully connected graph ${O}^{t} = (\mathcal{U}^t, \mathcal{E}^t)$, where $\mathcal{U}^t = \{\mathcal{D}_i^t, \mathcal{D}_j^t\}$ with all edges equal to 1. \\
    Mitigate adversarial noise on centroids via least-squares graph (Eq.~\ref{eq:LSMsol1}, \ref{eq:LSMsol2}), resulting in the set of resilient detections\\
    \hspace{1em} $\mathcal{J}_{ij}^t, \mathcal{J}_{ji}^t$ = Least-Squares Graph$\{\mathcal{D}_i^t, \mathcal{D}_j^t\}$\\
    \textbf{First Association Stage:} \\
    \hspace{1em} Associate $\mathcal{J}_{ij}^t$ with existing tracks $\mathcal{I}^t$: \\
    \hspace{2em} $m\mathcal{I}^t, m\mathcal{J}_{ij}^t, u\mathcal{I}^t, u\mathcal{J}_{ij}^t$ = Associate$\{\mathcal{J}_{ij}^t, \mathcal{I}^t\}$ \\
    \hspace{2em} Update, Initialize Tracks \\
    \textbf{Second Association Stage:} \\
    \hspace{1em} Associate $\mathcal{J}_{ji}^t$ with unmatched tracks $u\mathcal{I}^t$: \\
    \hspace{2em} $m\mathcal{I}^t, m\mathcal{J}_{ji}^t, u\mathcal{I}^t, u\mathcal{J}_{ji}^t$ = Associate$\{\mathcal{J}_{ji}^t, u\mathcal{I}^t\}$ \\
    \hspace{2em} Update, Initialize, Terminate Tracks \\
    Predict tracks for $t+1$ using Eq.~\ref{eq:prediction_sta}, \ref{eq:prediction_P}
}
\caption{\textbf{ARLOT Framework}}
\label{algor:ARLOT}
\end{minipage}
}
\end{algorithm}

\begin{figure} 
\centering
 \includegraphics[scale=0.6]{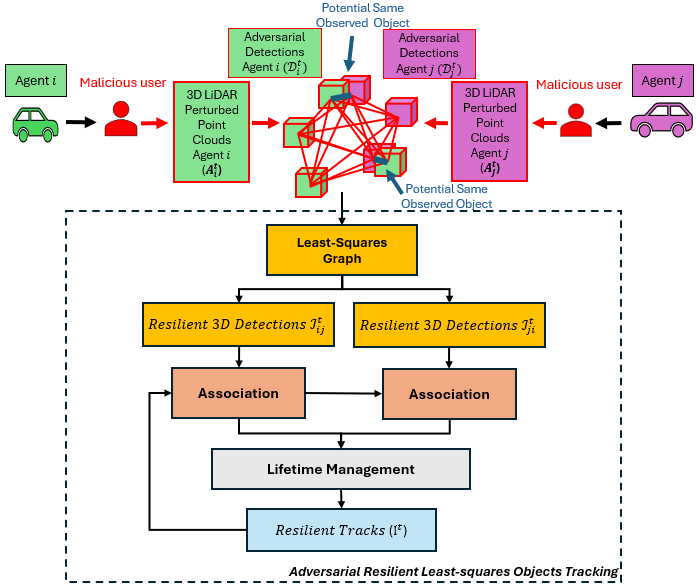}
  \caption{\textbf{ARLOT} approach}
  \label{fig:method1}
  \label{fig:method2}
\end{figure}

\section{Numerical Results}
\label{results}
\subsection{Experimental Setup and Metrics}
We have conducted extensive experiments on the real-world V2V4Real \cite{10203124} dataset of two driven vehicles, Tesla (Ego) and Astuff (Vehicle 1), under challenging adversarial conditions, using an NVIDIA RTX 4090 GPU.
Additionally, we define the adversarial parameters $k=40$ and $\alpha=\epsilon/30$, with $\epsilon=20cm$ or $25cm$ depending on the specific attack scenario and the tracking parameters $age$ = 2, and $hits$ = 3 to balance robustness and adaptability. We evaluated our framework on the testing sequences and compared its performance with the state-of-the-art baselines, including the V2V4Real and DMSTrack \cite{chiu2024probabilistic} pipelines for MAMOT and the AB3DMOT \cite{9341164} for SAMOT under adversarial attacks. Specifically, the V2V4Real approach uses multi-agent detections and employs the AB3DMOT for object tracking. The DMSTrack associates ego-vehicle's detections with existing trajectories and the unmatched trajectories are associated with those of the other vehicle. All methods employ the SECOND \cite{yan2018second} 3D object detector on each vehicle.
Furthermore, to enrich the insights of the resiliency of our method, we adopt a simple yet effective geometric transformation-based defense to mitigate the adversarial impact. Inspired by \cite{zhang2024comprehensive}, we apply a fixed clockwise rotation of $22.5^\circ$ around z-axis to each agent's adversarial point cloud.
To quantify the effect of the adversarial perturbation and rotation-based defense to point clouds on the inference of the detector, we employ the mean Average Precision, mAP. Moreover, we evaluate the resiliency of our proposed method based on the well-established evaluation tracking metrics of \cite{9341164} including 1) sAMOTA, 2) AMOTA, 3) AMOTP, 4) Mostly tracked (MT) (i.e., the percentage of correct tracking of objects over the 80\% of their life), with respect to True Positive (TP), False Positive (FP), False Negatives (FN), Identity Switches (IDSW) over the overlapping threshold of 0.25. 
\begin{table}[ht]
\centering
 \caption{Perception performance in terms of mAP, evaluated under benign and attacks conditions. Even with transformation based defense \cite{zhang2024comprehensive}, perception performance remains marginal.}  
\resizebox{6.5cm}{!}{
\begin{tabular}{|c|c|c|c|}
\hline
Vehicles & Noise & Without Defense (\%) $\uparrow$ & With Defense (\%) $\uparrow$\\
\hline
&benign & \textbf{48.83} &  -\\
Tesla & 20cm & 9.18 & 11.38\\      
& 25cm & 1.58 &  2.58 \\ 
\hline
&benign & \textbf{46.89} &  - \\
Astuff & 20cm & 14.1 & 17.55\\      
& 25cm & 5.01 &  7.79 \\  
\hline
\end{tabular}
}
\label{tab:mAP_detectors}
\end{table}
\begin{table}[ht]
\centering
\caption{Performance comparison on \textbf{Bening Point Clouds}. Plus (minus) sign indicate the rate of accuracy improvement (decrease) with respect to the maximum deviation of the state-of-the-art MAMOT methods.}\resizebox{8.9cm}{!}{ 
\begin{tabular}{|c|c|c|c|c|}
\hline
\textbf{Vehicles} & \textbf{sAMOTA (\%)$\uparrow$} & \textbf{AMOTA  (\%) $\uparrow$} & \textbf{AMOTP  (\%) $\uparrow$} & \textbf{MT (\%) $\uparrow$} \\
\hline
\textbf{Tesla \cite{9341164}} & 49.67 & 17.01 & 36.88 & 39.87 \\
\textbf{Astuff \cite{9341164}} & 56.79 & 21.11 & 37.03 & 45.10 \\
\textbf{V2V4Real \cite{10203124}} & 80.87 & 42.59 & 61.69 & 75.52 \\
\textbf{DMSTrack \cite{chiu2024probabilistic}} & \textbf{84.36} & \textbf{44.37} & 58.84 & 76.01\\
\textbf{ARLOT (ours)} & 83.45 (-1.08\%) & 42.85 (-3.43\%) & \textbf{65.39 (+11.13\%)} & \textbf{79.72 (+5.56\%)} \\
\hline
\end{tabular}
}
\label{tab:tracking_without_noise}
\end{table}
\begin{table}[ht]
\centering
\caption{Perception Performance on \textbf{Ego Vehicle} Under Attack.}
\resizebox{9cm}{!}{
\begin{tabular}{|c|c|c|c|c|c|}
\hline
\textbf{Noise}&\textbf{Tracking} & \textbf{sAMOTA (\%) $\uparrow$} & \textbf{AMOTA (\%) $\uparrow$} & \textbf{AMOTP (\%) $\uparrow$} & \textbf{MT (\%) $\uparrow$} \\
\hline 
&\textbf{Tesla \cite{9341164}} & 33.73 & 7.54 & 23.74 & 12.91 \\
Noise 20cm&\textbf{V2V4Real \cite{10203124}} & 74.74 & 36.25 & 50.58 & 56.36 \\
&\textbf{DMSTrack \cite{chiu2024probabilistic}} & 75.65 & 35.77 & 50.79 & 59.68\\
&\textbf{ARLOT (ours)} & \textbf{76.35 (+2.15\%)} & \textbf{38.06 (+6.4\%)} & \textbf{53.04 (+4.86\%)} & \textbf{ 61.84 (+9.72\%)} \\
\hline 
&\textbf{Tesla \cite{9341164}} & 36.83&8.78&26.82&16.67 \\
Noise 20cm+Defense&\textbf{V2V4Real \cite{10203124}}& 78.19&37.42&53.45&59.4\\
&\textbf{DMSTrack \cite{chiu2024probabilistic}} & 78.8 & 37.42 & 52.51 & 60.64\\
&\textbf{ARLOT (ours)}& \textbf{80.37 (+2.79\%)} & \textbf{40.3 (7.7\%)} & \textbf{56 (+6.65\%)} & \textbf{64.84 (+9.16\%)} \\
\hline 
&\textbf{Tesla \cite{9341164}} & 19.05&2.46&12.62&5.93 \\
Noise 25cm&\textbf{V2V4Real \cite{10203124}} & 66.78&29.23&43.28&50.75 \\
&\textbf{DMSTrack \cite{chiu2024probabilistic}} & 68.3 & 28.88 & 44.07 & \textbf{52.34}\\
&\textbf{ARLOT (ours)}& \textbf{70.46 (+5.51\%)} & \textbf{30.97 (+7.24\%)} & \textbf{46.92 (+8.41\%)} & \textbf{51.78 (-1.07\%)} \\
\hline 
&\textbf{Tesla \cite{9341164}} & 30.4& 2.77 & 13.63 & 4.88\\
Noise 25cm+Defense&\textbf{V2V4Real \cite{10203124}} & 66.95&30.24&43.71&51.17 \\
&\textbf{DMSTrack \cite{chiu2024probabilistic}} & \textbf{66.92} & 29.59 & 44.94 & \textbf{52.74}\\
&\textbf{ARLOT (ours)} & 67.96 (-1.79\%) & \textbf{31.66 (+7\%)} & \textbf{45.28 (+3.59\%)} & 50.9 (-3.48\%) \\
\hline
\end{tabular}
}
\label{tab:tracking_ego}
\end{table}

\begin{table}[ht]
\centering
\caption{Perception Performance on \textbf{Both Vehicles} Under Attack.}
\resizebox{9cm}{!}{
\begin{tabular}{|c|c|c|c|c|c|}
\hline
\textbf{Noise}&\textbf{Tracking} & \textbf{sAMOTA (\%) $\uparrow$} & \textbf{AMOTA (\%) $\uparrow$} & \textbf{AMOTP (\%) $\uparrow$} & \textbf{MT (\%) $\uparrow$} \\
\hline 
&\textbf{Tesla \cite{9341164}} & 33.73 & 7.54 & 23.74 & 12.91 \\
&\textbf{Astuff \cite{9341164}} &44.8&12.97&28.45&24.82\\
Noise 20cm&\textbf{V2V4Real \cite{10203124}} & 66.52&26.04&46.14&41.16 \\
&\textbf{DMSTrack \cite{chiu2024probabilistic}} & 69.27&27.92&45.24&41.6\\
&\textbf{ARLOT (ours)}& \textbf{69.85 (+5.02\%)} & \textbf{28.56 (+9.68\%)} & \textbf{49.03 (+8.38\%)} & \textbf{ 46.79 (+13.68\%)} \\
\hline 
&\textbf{Tesla \cite{9341164}} & 36.83&8.78&26.82&16.67 \\
&\textbf{Astuff \cite{9341164}}&43.73&13.1&27.92&27.81\\
Noise 20cm+Defense&\textbf{V2V4Real \cite{10203124}}& 66.19&26.93&46.9&45.3\\
&\textbf{DMSTrack \cite{chiu2024probabilistic}} & \textbf{68.23} & 28.9 & 45.5 & 44.28\\
&\textbf{ARLOT (ours)} & 67.87 (-0.53\%)& \textbf{29.37 (+9.06\%)} & \textbf{ 48.71 (+7.05\%)} & \textbf{49.85 (+12.58\%)} \\
\hline 
&\textbf{Tesla \cite{9341164}} & 19.05&2.46&12.62&5.93 \\
&\textbf{Astuff \cite{9341164}} &27.47&5.88&17.94&10.69\\
Noise 25cm&\textbf{V2V4Real \cite{10203124}}& 40.1&10.91&27.06&17.77 \\
&\textbf{DMSTrack \cite{chiu2024probabilistic}} & 42.18 & 11.84 & 27.42& \textbf{19.72}\\
&\textbf{ARLOT (ours)} & \textbf{46.07 (+14.91\%)} & \textbf{13.26 (+21.54\%)} & \textbf{30.53 (+12.82\%)} & 19.42 \textbf{(-1.52\%)} \\
\hline 
&\textbf{Tesla \cite{9341164}} & 30.4& 2.77 & 13.63 & 4.88\\
&\textbf{Astuff \cite{9341164}}&31.38&6.75&19.37&14.44\\
Noise 25cm+Defense&\textbf{V2V4Real \cite{10203124}} & 42.85&11.93&28.28&18.17 \\
&\textbf{DMSTrack \cite{chiu2024probabilistic}} & \textbf{48.22} & 13.96 &30.5 & 20.56\\
&\textbf{ARLOT (ours)} & 48.18 (-0.08\%) & \textbf{14.71 (+23.3\%)} & \textbf{32.22 (+13.93\%)} & \textbf{20.72 (+14.03\%)} \\
\hline
\end{tabular}
}
\label{tab:tracking_both}
\end{table}
\begin{figure}[ht]
\centering
 \includegraphics[scale=0.29]{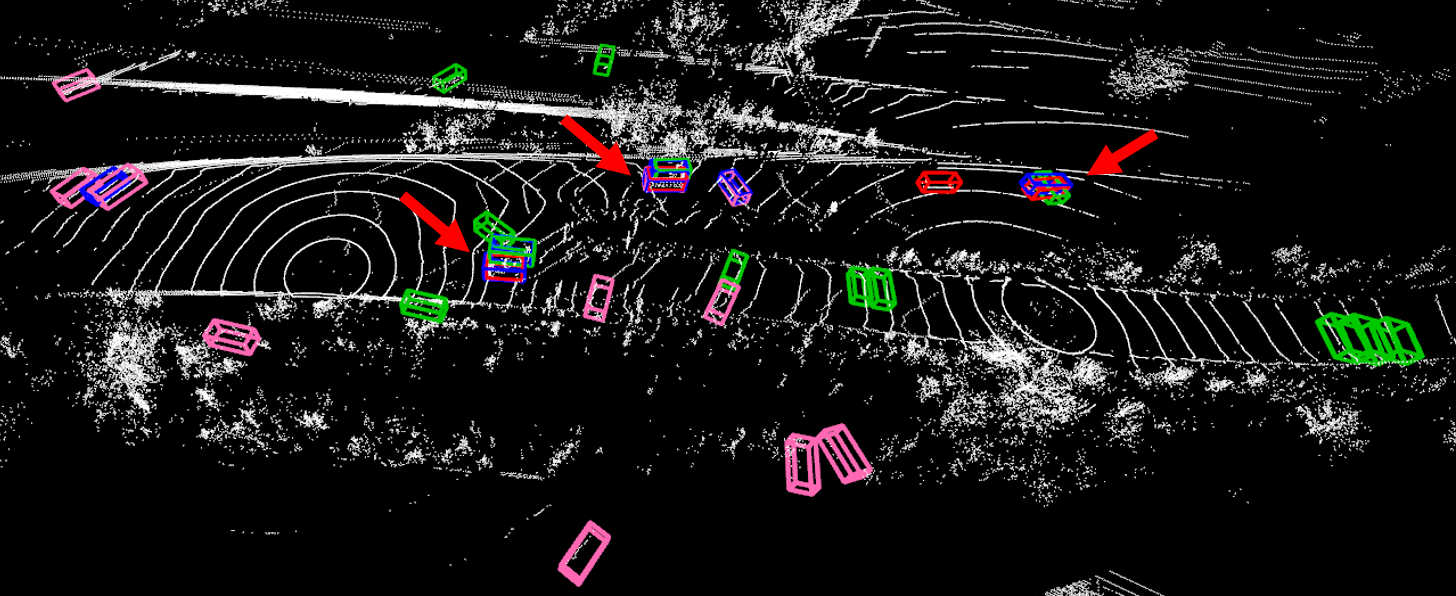}
  \caption{Sequence 0000 Frame 30. \textbf{Pink}: Ego detections. \textbf{Green}: Vehicle 1 detections. \textbf{Red}: Real Objects. \textbf{Blue}: \textbf{ARLOT} trajectories. \textbf{Red arrows} indicate the precise localization of objects under 20cm challenging adversarial noise on both agents.}
  \label{fig:visual_20}
\end{figure}
\subsection{Evaluation Study}
Table \ref{tab:mAP_detectors} demonstrates the impact of adversarial noise on individual detection performance for two CAVs, as well as the mitigation achieved through rotation-based defense. 
Introducing 20cm perturbations to point clouds significantly increases misdetections and perturbed detections with a maximum mAP dropping by up to \textbf{81.2\%}, severely degrading detection performance. At 25cm noise, the attack has even stronger effect, resulting in marginal perception. Although, rotation defense enhances accuracy by mitigating the adversarial noise, overall perception performance is still. Hence, further investigation on perception robustness is required, and especially how MAMOT can actually alleviate the adversarial attacks impact.

Table \ref{tab:tracking_without_noise} presents the tracking performance of the proposed, and the baseline methods on average across all testing sequences. The \textbf{ARLOT} achieves superior performance, with the maximum \textbf{11.13\%} and \textbf{77.3\%} improvements in AMOTP on MAMOT and SAMOT respectively, localizing the objects precisely in the scene reducing the positional error of multi-agent detections through the least-squares graph. 
Furthermore, its tracking consistency achieves maximum scores by \textbf{5.56\%} and \textbf{99.95\%} instead of the baseline state-of-the-art MAMOT and SAMOT approaches. Although \textbf{ARLOT} improves tracking accuracy over SAMOT by up to \textbf{151.91\%}, the sequential DMSTrack achieves better results.
Hence, the proposed method attains promising results in object localization, as well as effective associations, by smoothing the detections' spatial errors via the least-squares graph with the overlapped data, motivating us to apply our framework as a denoiser against adversarial noise,
especially under severe conditions.

Table \ref{tab:tracking_ego} presents the perception performance on tracking metrics under two levels of adversarial noise on the ego vehicle’s point clouds, both with and without rotation defense. At 20cm adversarial noise, the \textbf{ARLOT} achieves notable improvements of \textbf{2.15\%} in sAMOTA and \textbf{6.4\%} in AMOTA over the MAMOT baselines, and similar results with rotation applied.
Hence, \textbf{ARLOT} effectively tracks more TP and fewer FN objects in the scene, via its two-stage association maintaining robust performance without external defense strategies. Furthermore, \textbf{ARLOT} outperforms in AMOTP by \textbf{4.86\%} and \textbf{9.72\%} in MT without rotation. 
Therefore, our MAMOT effectively mitigates adversarial centroid errors in multi-agent detections using least-squares graph with the differential coordinates and the anchor points. Notably, \textbf{ARLOT} achieves \textbf{404.77\%} higher accuracy, and \textbf{123.42\%} precision compared to SAMOT, highlighting the critical role of agent cooperation under adversarial conditions. Similar results are presented under the challenging adversarial noise of 25cm, indicating that the \textbf{ARLOT}
achieves resilience through the least-squares graph exploiting the multi-agent adversarial information.   

To study further the resilience of \textbf{ARLOT} framework under severe adversarial attacks, Table 
\ref{tab:tracking_both} demonstrates the perception performance under attacking to both CAVs, with and without rotation-based defense. At 25cm adversarial noise, \textbf{ARLOT} achieves the highest scores in sAMOTA by \textbf{14.91\%} and in AMOTA by \textbf{21.54\%} than the MAMOT baselines by reducing FN and FP errors, and increasing the TP. These results indicate the effective smoothness of multi-agent adversarial detections via differential coordinates, anchor points and two-stages tracking association. Moreover, the \textbf{ARLOT} outperforms with maximum improvements up to \textbf{12.82\%} and \textbf{141.92\%} in AMOTP compared to baseline MAMOT and SAMOT pipelines, confirming its resilience on reducing the adversarial noise via the least-squares graph scheme.
Hence, \textbf{ARLOT} outperforms state-of-the-art MAMOT and SAMOT pipelines in all metrics under challenging adversarial noise, with or without defense mechanisms. The proposed Graph based method smoooths the displaced multi-agent detections reducing accuracy errors and serving as a resilient scheme requiring only sharing detections without training.

To provide deeper insight into the benefits of \textbf{ARLOT} as a resilient method in perception, Fig.\ref{fig:visual_20} illustrates the detections of the ego (pink color) and vehicle 1 (green color) along with the GT object (red color) and the trajectories of \textbf{ARLOT}, under 20cm adversarial noise on both CAVs. As highlighted by the red arrows, \textbf{ARLOT} maintains accurate and precise object trajectories, showcasing its ability to effectively mitigate adversarial impact even under severe adversarial conditions.

\section{Conclusion}
\label{conclusion}
In this paper, a novel MAMOT method, resilient against adversarial attacks is proposed, exploiting the multi-vehicle adversarial detections. The proposed so called \textbf{ARLOT} approach formulates a fully connected graph of predicted bounding boxes and via Laplacian processing reduces their induced spatial error. 
The optimized detections update tracks' states via Kalman Filters in two stages, achieving precise localization and identification of 3D objects, operating further as a denoiser against adversarial perturbations. Hence, integrating the least-squares graph framework within the overall MAMOT concept for increased and resilient perception, our \textbf{ARLOT} effectively addresses adversarial attacks perturbations, outperforming state-of-the-art SAMOT and MAMOT methods by up to \textbf{404.77\%} and \textbf{23.3\%} respectively, without additional defense mechanisms. Future work will involve the investigation of robustness of our method on multi-X real-world sensor data.


\begin{thebibliography}{10}
\providecommand{\url}[1]{#1}
\csname url@samestyle\endcsname
\providecommand{\newblock}{\relax}
\providecommand{\bibinfo}[2]{#2}
\providecommand{\BIBentrySTDinterwordspacing}{\spaceskip=0pt\relax}
\providecommand{\BIBentryALTinterwordstretchfactor}{4}
\providecommand{\BIBentryALTinterwordspacing}{\spaceskip=\fontdimen2\font plus
\BIBentryALTinterwordstretchfactor\fontdimen3\font minus \fontdimen4\font\relax}
\providecommand{\BIBforeignlanguage}[2]{{%
\expandafter\ifx\csname l@#1\endcsname\relax
\typeout{** WARNING: IEEEtran.bst: No hyphenation pattern has been}%
\typeout{** loaded for the language `#1'. Using the pattern for}%
\typeout{** the default language instead.}%
\else
\language=\csname l@#1\endcsname
\fi
#2}}
\providecommand{\BIBdecl}{\relax}
\BIBdecl

\bibitem{szegedy2013intriguing}
C.~Szegedy, \textit{et al.}, ``Intriguing properties of neural networks,'' \emph{arXiv preprint arXiv:1312.6199}, 2013.

\bibitem{xiang2019generating}
C.~Xiang, \textit{et al.}, ``Generating 3d adversarial point clouds,'' in \emph{Proceedings of the IEEE/CVF conference on computer vision and pattern recognition}, 2019, pp. 9136--9144.

\bibitem{liu2020adversarial}
D.~Liu, \textit{et al.}, ``Adversarial shape perturbations on 3d point clouds,'' in \emph{Computer Vision--ECCV 2020 Workshops: Glasgow, UK, August 23--28, 2020, Proceedings, Part I 16}.\hskip 1em plus 0.5em minus 0.4em\relax Springer, 2020, pp. 88--104.

\bibitem{wicker2019robustness}
M.~Wicker, \textit{et al.}, ``Robustness of 3d deep learning in an adversarial setting,'' in \emph{Proceedings of the IEEE/CVF Conference on Computer Vision and Pattern Recognition}, 2019, pp. 11\,767--11\,775.

\bibitem{zhang2024comprehensive}
Y.~Zhang, \textit{et al.}, ``A comprehensive study of the robustness for lidar-based 3d object detectors against adversarial attacks,'' \emph{International Journal of Computer Vision}, vol. 132, no.~5, pp. 1592--1624, 2024.

\bibitem{jia2024robust}
S.~Jia, \textit{et al.}, ``Robust deep object tracking against adversarial attacks,'' \emph{International Journal of Computer Vision}, pp. 1--20, 2024.

\bibitem{zhang2024dual}
W.~Zhang, \textit{et al.}, ``Dual-dimensional adversarial attacks: A novel spatial and temporal attack strategy for multi-object tracking,'' in \emph{2024 International Joint Conference on Neural Networks (IJCNN)}.\hskip 1em plus 0.5em minus 0.4em\relax IEEE, 2024.

\bibitem{jia2020fooling}
Y.~Jia, \textit{et al.}, ``Fooling detection alone is not enough: Adversarial attack against multiple object tracking,'' in \emph{International Conference on Learning Representations (ICLR'20)}, 2020.

\bibitem{long2024papmot}
J.~Long, \textit{et al.}, ``Papmot: Exploring adversarial patch attack against multiple object tracking,'' in \emph{European Conference on Computer Vision}.\hskip 1em plus 0.5em minus 0.4em\relax Springer, 2024, pp. 128--144.

\bibitem{muller2022physical}
R.~Muller, \textit{et al.}, ``Physical hijacking attacks against object trackers,'' in \emph{Proceedings of the 2022 ACM SIGSAC Conference on Computer and Communications Security}, 2022, pp. 2309--2322.

\bibitem{shuai2021siammot}
B.~Shuai, \textit{et al.}, ``Siammot: Siamese multi-object tracking,'' in \emph{Proceedings of the IEEE/CVF conference on computer vision and pattern recognition}, 2021, pp. 12\,372--12\,382.

\bibitem{tu2021adversarial}
J.~Tu, \textit{et al.}, ``Adversarial attacks on multi-agent communication,'' in \emph{Proceedings of the IEEE/CVF International Conference on Computer Vision}, 2021, pp. 7768--7777.

\bibitem{wu2024enhancing}
Z.~Wu, \textit{et al.}, ``Enhancing tracking robustness with auxiliary adversarial defense networks,'' in \emph{European Conference on Computer Vision}.\hskip 1em plus 0.5em minus 0.4em\relax Springer, 2024, pp. 198--214.

\bibitem{yang2019adversarial}
J.~Yang, \textit{et al.}, ``Adversarial attack and defense on point sets,'' \emph{arXiv preprint arXiv:1902.10899}, 2019.

\bibitem{zhao2024made}
Y.~Zhao, \textit{et al.}, ``Made: Malicious agent detection for robust multi-agent collaborative perception,'' in \emph{2024 IEEE/RSJ International Conference on Intelligent Robots and Systems (IROS)}.\hskip 1em plus 0.5em minus 0.4em\relax IEEE, 2024, p. 13817.

\bibitem{ding2023robust}
S.~Ding, \textit{et al.}, ``Robust multi-agent communication with graph information bottleneck optimization,'' \emph{IEEE Transactions on Pattern Analysis and Machine Intelligence}, vol.~46, no.~5, pp. 3096--3107, 2023.

\bibitem{sorkine2004least}
O.~Sorkine, \textit{et. al.}, ``Least-squares meshes,'' in \emph{Proceedings Shape Modeling Applications, 2004.}\hskip 1em plus 0.5em minus 0.4em\relax IEEE, 2004, pp. 191--199.

\bibitem{10203124}
R.~Xu, \textit{et al.}, ``V2v4real: A real-world large-scale dataset for vehicle-to-vehicle cooperative perception,'' in \emph{2023 IEEE/CVF Conference on Computer Vision and Pattern Recognition (CVPR)}, 2023.

\bibitem{9341164}
X.~Weng, \textit{et al.}, ``3d multi-object tracking: A baseline and new evaluation metrics,'' in \emph{2020 IEEE/RSJ International Conference on Intelligent Robots and Systems (IROS)}, 2020, pp. 10\,359--10\,366.

\bibitem{piperigkos2024graph}
N.~Piperigkos, \textit{et al.}, ``Graph laplacian processing based multi-modal localization backend for robots and autonomous systems,'' \emph{IEEE Transactions on Cognitive and Developmental Systems}, 2024.

\bibitem{chiu2024probabilistic}
H.-K. Chiu, \textit{et al.}, ``Probabilistic 3d multi-object cooperative tracking for autonomous driving via differentiable multi-sensor kalman filter,'' in \emph{2024 IEEE International Conference on Robotics and Automation (ICRA)}.\hskip 1em plus 0.5em minus 0.4em\relax IEEE, 2024, pp. 18\,458--18\,464.

\bibitem{yan2018second}
Y.~Yan, \textit{et al.}, ``Second: Sparsely embedded convolutional detection,'' \emph{Sensors}, vol.~18, no.~10, p. 3337, 2018.

\end{thebibliography}
\end{document}